\documentclass[letterpaper, 10 pt, conference]{ieeeconf}  

\IEEEoverridecommandlockouts                              

\usepackage{graphics}
\usepackage{epsfig}
\usepackage{times}

\usepackage{booktabs}
\usepackage{tikz}
\usetikzlibrary{automata,fit,positioning,arrows,backgrounds}

\usepackage{pgfplotstable}
\pgfplotsset{compat=1.15}

\title{\LARGE \bf
    Fine-grained activity recognition for assembly videos
}

\author{Jonathan D. Jones$^{1}$, %
Cathryn Cortesa$^{2}$, %
Amy Shelton$^{3}$, %
Barbara Landau$^{2}$, %
\\ %
Sanjeev Khudanpur$^{1}$, %
and %
Gregory D. Hager$^{4}$
\thanks{This work was supported by NSF award Nos. 1561278 and IIS-1900952.}
\thanks{$^{1}$Jonathan D. Jones and Sanjeev Khudanpur are with the Department of Electrical Engineering, Johns Hopkins University, Baltimore, MD, USA {\tt\small jdjones@jhu.edu}}%
\thanks{$^{2}$Cathryn Cortesa and Barbara Landau are with the Department of Cognitive Science, Johns Hopkins University, Baltimore, MD, USA}%
\thanks{$^{3}$Amy Shelton is with the School of Education, Johns Hopkins University, Baltimore, MD, USA}%
\thanks{$^{4}$Gregory D. Hager is with the Department of Computer Science, Johns Hopkins University, Baltimore, MD, USA}%
}

\usepackage{xifthen}
\usepackage{xcolor}

\newcommand{\ie}{\emph{i.e.}~}
\newcommand{\eg}{\emph{e.g.}~}

\usepackage{amsmath}
\usepackage{amssymb}
\usepackage{amsthm}
\usepackage{mathtools}

\DeclarePairedDelimiter{\norm}{\lVert}{\rVert}
\newcommand{\parens}[1]{\left( #1 \right)}

\newcommand{\angles}[1]{\left\langle #1 \right\rangle}
\newcommand{\braces}[1]{\left\{ #1 \right\}}

\DeclareMathOperator*{\argmax}{arg\,max}

\DeclareMathOperator{\pOp}{P}

\newcommand{\CondProb}[2]{\pOp \! \parens{ #1 \mid #2 }}

\newcommand{\vertices}{\mathcal{V}}
\newcommand{\edges}{\mathcal{E}}

\begin{document}

\maketitle
\thispagestyle{empty}
\pagestyle{empty}

\begin{abstract}
In this paper we address the task of recognizing assembly actions as a structure (\eg a piece of furniture or a toy block tower) is built up from a set of primitive objects.
Recognizing the full range of assembly actions requires perception at a level of spatial detail that has not been attempted in the action recognition literature to date.
We extend the fine-grained activity recognition setting to address the task of assembly action recognition in its full generality by unifying assembly actions and kinematic structures within a single framework.
We use this framework to develop a general method for recognizing assembly actions from observation sequences,
along with observation features that take advantage of a spatial assembly's special structure.
Finally, we evaluate our method empirically on two application-driven data sources:
(1) An IKEA furniture-assembly dataset, and (2) A block-building dataset.
On the first, our system recognizes assembly actions with an average framewise accuracy of 70\% and an average normalized edit distance of 10\%.
On the second, which requires fine-grained geometric reasoning to distinguish between assemblies,
our system attains an average normalized edit distance of 23\%---a relative improvement of 69\% over prior work.
\end{abstract}

\section{Introduction}
\label{sec:introduction}
Motivated by a desire to enable better human-robot collaboration and finer-grained behavioral analyses, researchers in computer vision and robotics have recently begun to approach the challenging problem of assembly action recognition \cite{action-gupta-uist-2012, hadfield-blocks-iros-2018, robots-wang-hri-2020, blocks-jones-wacv-2019, costar-blocks}. In assembly activity recognition, a perception system must recognize {\em both} the assembly actions {\em and} the configuration of a structure (\eg a piece of furniture or a toy block tower) as it is built up from a set of primitive objects.


Recognizing assembly actions requires perception at a finer level of detail than most activity recognition research to date.
Typical action recognition methods are concerned with modeling scene-level information, and 
usually involve the classification of a very large set of short video clips, each on the order of tens of seconds to a minute, drawn from a diverse set of high-level action categories.
For example, the benchmark dataset UCF101 entails recognizing actions like \texttt{playing guitar}, \texttt{playing piano}, \texttt{cutting in kitchen}, and \texttt{skydiving}.
In \emph{fine-grained} action recognition, on the other hand, there is less inter-class variability between categories. 
For instance, a system that automatically generates a recipe from a cooking video will need to distinguish between action categories defined at a finer level than \texttt{cutting in kitchen}---it needs to understand \emph{what} is being cut, and \emph{when}.
These approaches are generally concerned with perception at the level of objects, rather than scenes.

Recognizing assembly actions requires perception at a level of geometric detail even finer than object-level, which, to our knowledge, has not been attempted in the action recognition literature to date.
Systems must perceive not only \emph{which} objects are connected to each other and \emph{when}, but also \emph{where} they are connected---that is, which specific contact points (screw holes, for example) are connected between two parts. In short, assembly action recognition requires a framework that unifies both semantic and geometric computer vision.

In this paper we develop and demonstrate a system that is able to perform assembly action recognition. In doing so we make three contributions:
\begin{enumerate}
    \item We develop an assembly action recognition framework that unifies  previous approaches in activity recognition with the notion of a {\em kinematic state}, and relate the two by defining an action as a difference between two kinematic states.
\item We outline a general method for recognizing assembly actions from observation sequences, along with features that take advantage of a spatial assembly's special structure.
\item Finally, we evaluate our method empirically on two challenging real-world data sets:  egocentric videos of furniture assembly, and third-person RGB-D video and inertial measurements recorded during toy-block assembly.
\end{enumerate}

\section{Related work}
\label{sec:related_work}

\subsection{Activity recognition}
Work in activity recognition has focused on the temporal structure of assembly. To date, approaches have addressed a simplified version of the assembly setting in which there is at most one way to connect each pair of parts and parts are never disconnected. In other words, they treat assembly actions at the object level, rather than at the level of intra-object contact points. This prevents systems from recognizing the full range of assembly actions that may be encountered in realistic deployment scenarios.

Summers-Stay and colleagues applied an action grammar to the task of parsing a handful of kitchen and craft activities---namely 
\texttt{cooking vegetables}, 
\texttt{making sandwich}, 
\texttt{sewing a toy}, 
\texttt{card making}, and
\texttt{assemble a machine} \cite{action-summers-iros-2012}.
This work was part of a larger effort to develop a formal grammar on manipulation actions in analogy to the Minimalist Program (an influential framework in the field of theoretical syntax) 
\cite{action-yang-acl-2015}.
Central to their approach is the MERGE operator, which they use to represent actions and compute their effects. They observed that just as a noun and a determiner can be combined to form a noun phrase, distinct object instances are combined into composite entities during manipulation actions. However, where the non-terminal nodes in a syntactic parse tree represent interpretable linguistic concepts, the non-terminals in their activity trees simply represent the order in which objects were combined. Additionally, this grammar has no mechanism for handling the disassembly of an object---they note that future work should study ``the ability to build up trees based on disassembly and transformation as well as assembly''. 

A similar effort by Vo and Bobick
\cite{action-vo-cviu-2016}
used hand-defined probabilistic context-free grammars (PCFGs) to instantiate a graphical model for the purpose of segmenting and classifying compositional action sequences.
They evaluated their method on videos of toy airplane construction, along with cooking and human activity datasets.
Like the work of Summers-Stay and colleagues, this system produces from an assembly video a parse tree whose terminal nodes represent primitive objects and whose non-terminals represent composite entities (\ie partial assemblies of the final product). However, this work additionally required that non-terminals correspond to interpretable sub-parts, such as the nose, wings, body, and tail of an airplane. While this requirement suits the hierarchical structure of a PCFG, it may be unrealistic for actual assembly parsing applications---any build sequence that does not conform to the pre-specified sequence of sub-parts could not be recognized by their model, even if it resulted in a final assembly that was present in the dataset.
For instance, their framework has no way to represent an assembly sequence in which someone builds the body,
then builds part of the nose, then builds the tail, and finally comes back to finish the nose.

Both methods ultimately fall short in actual assembly parsing applications because they over-simplify their representation of assembly actions.
In these situations, representing the fine-grained detail of part connection (\eg how and where parts are connected to each other) and allowing for disconnection actions is more important than a sequence model with recursive hierarchical structure.
By relating action sequences to an assembly's kinematic structure,
our representation captures the full variability assembly situations.

\subsection{Robotic Perception}
The robotic perception literature has focused on kinematic rather than temporal structure.
Although prior work exists going back to the late '90s 
\cite{kinem-models-costeira-ijcv-1998},
the first complete probabilistic model for identifying kinematic systems was developed by Sturm and colleagues
\cite{kinem-models-sturm-jair-2011},
Using a graphical model to account for probabilistic dependencies between objects,
this method estimated a time-invariant kinematic system
from position data obtained using fiducial markers and, in one experiment,
from plane-fitting on range data.
Later, Niekum and colleagues \cite{kinem-models-niekum-icra-2015} extended this model to perform changepoint detection of time-varying
kinematic systems.
Using their method, they identified configurations which can alter
the kinematic behavior of a system (for example, the angle at which a stapler's hinge locks in place).
In a related line of work, Mart\'{i}n-Mart\'{i}n and Brock  used a hierarchy of extended Kalman filters
to estimate static kinematic systems by tracking features in an RGB-D video stream
\cite{%
kinem-models-martin-ijrr-2019%
}.
They also extended their method to multiple modalities by including a robot's end-effector motion as
another observation signal.

These efforts have produced a rigorous and detailed theoretical framework
for the estimation of kinematic structures from a variety of pose or location data
(\ie object poses, feature point locations, or end-effector configurations).
However, even methods that analyze time-varying kinematic systems currently lack
a framework for describing systems that evolve dynamically---for example, due to the effects of assembly actions.

\section{Spatial assembly sequences}
\label{sec:state_machine}
%


Previous approaches to recognizing assembly actions (or equivalently, assembly structures) are scattered across several sub-disciplines in vision and robotics, and each addresses its own task independently. However, in this paper we observe that there is a fundamental structure to the problem that is shared by all methods. In the following sections we develop and evaluate a framework that unifies approaches based on action recognition and approaches based on kinematic model estimation. We treat an assembly process as a sequence of kinematic structures which transition from one to another as the result of a user's action. Then, we use our model to recognize assembly actions during a furniture construction task \cite{robots-wang-hri-2020} and a toy block building task \cite{blocks-jones-wacv-2019}.


\subsection{Spatial assemblies as kinematic constraint graphs}
\label{sec:assembly_graphs}

\begin{figure}[t]
    \centering
    \includegraphics[scale=0.25]{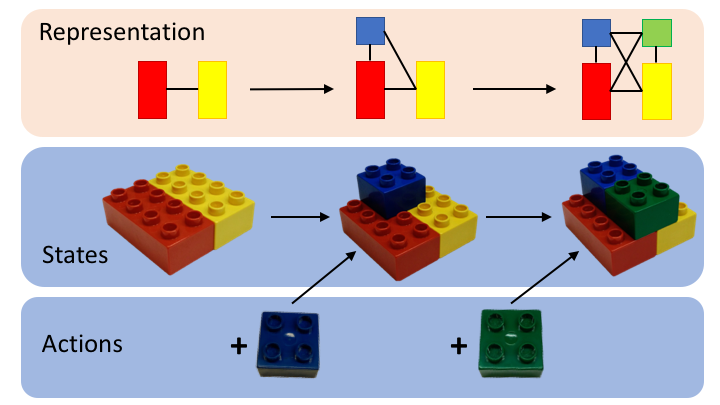}
    \caption{
        Partial example of an assembly process.
        The agent alters the state by adding or removing blocks in specific ways.
    }
    \label{fig:assembly-seq}
\end{figure}

Consider as an example the assembly sequence shown in Figure \ref{fig:assembly-seq},
in which a person begins with two rectangular blocks aligned along their longest side and consecutively adds two square blocks.
Each action the builder takes causes the assembly to transition to a new state,
with a different set of kinematic joints.

Generalizing this example, we represent a spatial assembly as a graph whose vertices correspond to primitive objects (\ie its component parts). If two objects are joined to each other, their corresponding vertices are connected by an edge representing the constraint imposed by that joint.
Each vertex is associated with attributes related to the geometric structure of the part: these attributes identify the types of contact points on the part and their locations.
Likewise, each edge is associated with attributes that specify which contact points are joined between two connected parts. In what follows we will refer to the vertex set  as $\vertices = \braces{v_1, \ldots, v_n}$, and the edge set as $\edges = \braces{ e_1, \ldots, e_m }$.


In some situations we will encounter assemblies whose objects form multiple, disjoint sub-parts.
In terms of our graph representation,
each sub-part corresponds to a separate connected component in the graph.

\subsection{Assembly sequences}
A builder assembles an object over time by taking actions that change that object's kinematic structure.
In general, kinematic changes can take many forms: for example,
many folding mechanisms such as chairs or gates are capable of locking into a stable configuration with more degrees of freedom.
In assembly processes, these kinematic changes take the form of
elementary actions that incorporate parts into a structure or remove them from it.
Previous assembly action recognition representations were essentially of the form
$\mathtt{add(part)}$ \cite{action-vo-cviu-2016} or $\mathtt{MERGE(part_1, part_2)}$ \cite{action-summers-iros-2012}.
We extend this representation to apply to general assembly scenarios by observing that an assembly action can be thought of as a difference between two states: the current configuration of a spatial assembly and the new configuration that is produced by the action.

Our action representation is comprised of an action type ($\texttt{connect}$ or $\texttt{disconnect}$)
and an assembly update $\delta s$.
$\delta s$ is a \emph{partial} kinematic graph: it has the same edge, vertex, and attribute structure we defined in Section \ref{sec:assembly_graphs}, but it describes changes in relationships between links rather than the complete kinematic state.
With this information, we can compute the result of taking any action $a$, given the current configuration of the assembly $s$:
\begin{itemize}
\item
$
    \texttt{connect}(s, \delta s) \coloneqq s \cup \delta s
$
\item
$
    \texttt{disconnect}(s, \delta s) \coloneqq s \setminus \delta s.
$
\end{itemize}
Familiar assembly actions an be implemented as instances of these action types: for example, \texttt{screw}/\texttt{unscrew} and \texttt{stack}/\texttt{remove} actions may impose different kinematic constraints $\delta s$.
In this way our two basic action types behave like the sign of an assembly action.

A spatial assembly sequence is formed as a builder takes actions over time.
The assembly begins in an initial configuration $s_0$,
and each action $a_1, a_2, \ldots, a_M$ produces a new configuration $s_1, \ldots, s_M$.
Although an assembly sequence can in principle begin in any state,
in our experiments we will always begin at the state in which no objects are connected
(we call this the \emph{empty} state, because its edge set is empty).

\section{Recognizing assembly actions}
\label{sec:model}

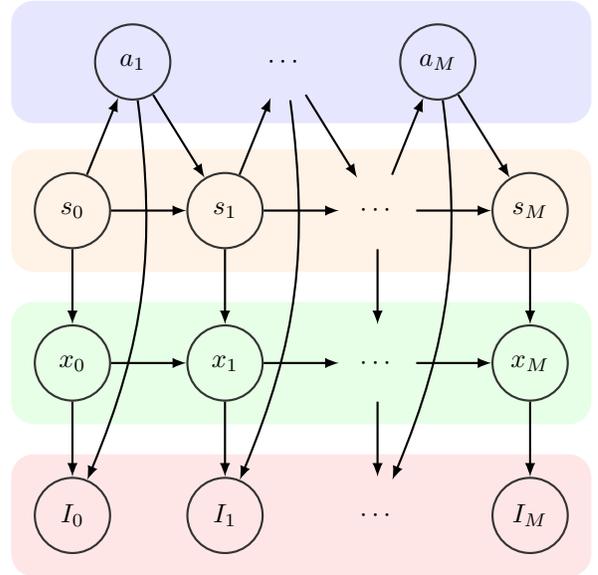
\begin{figure}[t]
    \centering
    \begin{tikzpicture}
    \tikzstyle{main}=[circle, minimum size = 10mm, thick, draw = black!80, node distance = 10mm]
    \tikzstyle{connect}=[-latex, thick]
    \tikzstyle{box}=[rectangle, draw=black!100]
    \tikzstyle{blank}=[main, draw=none, fill=none]
    
    
      \node[main] (s1) {$s_0$};
      \node[main] (s2) [right=of s1] {$s_1$};
      \node[blank] (s3) [right=of s2] {$\cdots$};
      \node[main] (s4) [right=of s3] {$s_M$};
      \node[main] (a1) [above left=12.5mm and 5mm of s2] {$a_1$};
      \node[blank] (a2) [above left=12.5mm and 5mm of s3] {$\cdots$};
      \node[main] (a3) [above left=12.5mm and 5mm of s4] {$a_M$};
      \node[main] (x1) [below=of s1] {$x_0$};
      \node[main] (x2) [below=of s2] {$x_1$};
      \node[blank] (x3) [below=of s3] {$\cdots$};
      \node[main] (x4) [below=of s4] {$x_M$};
      \node[main] (y1) [below=of x1] {$I_0$};
      \node[main] (y2) [below=of x2] {$I_1$};
      \node[blank] (y3) [below=of x3] {$\cdots$};
      \node[main] (y4) [below=of x4] {$I_M$};
      \path (s1) edge [connect] (s2)
            (s2) edge [connect] (s3)
            (s3) edge [connect] (s4);
      \path (x1) edge [connect] (x2)
            (x2) edge [connect] (x3)
            (x3) edge [connect] (x4);
      \path (s1) edge [connect] (a1)
            (a1) edge [connect] (s2)
            (s2) edge [connect] (a2)
            (a2) edge [connect] (s3)
            (s3) edge [connect] (a3)
            (a3) edge [connect] (s4);
      \path (s1) edge [connect] (x1)
            (x1) edge [connect] (y1);
      \path (s2) edge [connect] (x2)
            (x2) edge [connect] (y2);
      \path (s3) edge [connect] (x3)
            (x3) edge [connect] (y3);
      \path (s4) edge [connect] (x4)
            (x4) edge [connect] (y4);
      \path (a1) edge [connect, bend left=15] (y1);
      \path (a2) edge [connect, bend left=15] (y2);
      \path (a3) edge [connect, bend left=15] (y3);
    
    \begin{pgfonlayer}{background}
        \filldraw [line width=6mm,join=round,blue!10]
          (a1.north  -| s1.west)  rectangle (a3.south  -| s4.east);
        \filldraw [line width=6mm,join=round,orange!10]
          (s1.north  -| s1.west)  rectangle (s4.south  -| s4.east);
        \filldraw [line width=6mm,join=round,green!10]
          (x1.north  -| x1.west)  rectangle (x4.south  -| x4.east);
        \filldraw [line width=6mm,join=round,red!10]
          (y1.north  -| y1.west)  rectangle (y4.south  -| y4.east);
    \end{pgfonlayer}
\end{tikzpicture}
    \caption{
        Graphical model depicting the generation of a video sequence from an assembly process. The shaded regions depict the various processing stages into which the task is frequently decomposed.
        From bottom to top: object recognition, object tracking, kinematic perception, action recognition.
        Each variable in the figure corresponds to a segment of the video---\ie variables indexed with 0 correspond to the video segment depicting the assembly in its initial state and so on.
    }
    \label{fig:parser-pgm}
\end{figure}

In the assembly action recognition setting, we observe a sequence of discrete-time samples $y = y_1, \ldots, y_T$. Here $y$ usually represents a video, but could also be a sequence of object poses or mid-level attribute predictions.
Each individual assembly $s_m$ is realized as a segment of these samples: the $m$-th assembly generates the observation segment $y_{b_m}, \ldots, y_{e_m}$, where $b_m$ is the beginning index of segment $m$ and $e_m$ is its ending index.
In a slight abuse of notation, we will use $s = s_1, \ldots, s_T$ to refer to the assembly labels for each individual sample $1, \ldots, T$.
The assembly action recognition task consists of segmenting and classifying an observation sequence, producing as output a sequence of assembly structures $s = s_1, \ldots, s_T$ and/or a corresponding sequence of assembly actions $a = a_1, \ldots a_T$.
In Figure \ref{fig:parser-pgm} we illustrate a graphical model which depicts the generation of a video sequence from an assembly process\footnote{In defining this generative process we are not prescribing that all approaches to assembly action recognition be strictly generative, or even probabilistic---rather, we use it as a tool to examine the problem and compare methods.}.

\subsection{A segmental CRF for assembly action recognition}
\label{sec:model-fst}
%
We implement our models using a segmental conditional random field (CRF). This family of models is useful in the assembly action recognition setting because they generalize both traditional probabilistic methods (\eg hidden Markov models or their segmental generalizations) and contemporary neural methods (complex, learned features that feed into a logistic classifier) \cite{crfs}.
This provides the flexibility to to implement first-principles, probabilistic methods for applications with few to no examples to train on, to incorporate statistically-learned classifiers (\eg neural networks) when datasets are big enough that learning can improve performance, to enforce temporal consistency in the model's output.

\subsubsection{Scoring assembly observations}
We score each individual segment using a weighted combination of observation features $f_{obs}$. These features which model the relationship between observations and kinematic structures (or actions), and usually include object poses as an auxiliary variable:
\begin{align} \label{eq:score_feats}
\begin{split}
    score \parens{y, s_m, s_{m + 1}} &=
        \sum_{t} W_{obs}^T f_{obs} \parens{y_t, s_m, s_{m + 1}}
\end{split}
\end{align}

Our observation features score state \emph{transitions}: they can depend on both the current assembly structure $s_m$ and the next assembly $s_{m+1}$. This allows us to score action-related evidence and assembly-related evidence using the same framework---for example, identifying a screwdriver in use is a good indication that a \texttt{screwing} or \texttt{unscrewing} action is happening at that instant. On the other hand, observing that two parts move with a constant relative pose means they are probably connected on the same structure. For action features,
$f_{obs} \parens{y_t, s_m, s_{m+1}} = f(y_t, s_{m+1} - s_m) = f(y_t, a_{m+1})$.
For assembly features,
$f_{obs} \parens{y_t, s_m, s_{m+1}} = f(y_t, s_m)$.
The exact observation features we use will depend on the characteristics of the application data---for example, we use distances between estimated object poses for a furniture-assembly task with AR markers,
and we use pixel-level template registration scores along with IMU-based attributes for the block-play dataset.

We can take advantage of a spatial assembly's structure to design part-based assembly features. These features compare mid-level attributes individually for each pair of vertices in the graph:
\begin{align} \label{eq:obs_feat}
\begin{split}
    f_{obs} \parens{y_t, s_m} &= \sum_{i=1}^{|\mathcal{V}|} \sum_{j=1}^{i-1}
        g \parens{ \hat{a}_{ij}\parens{y_t}, a_{ij} \parens{s_m}}
\end{split}
\end{align}
$\hat{a}_{ij}\parens{y_t}$ represents the estimated value of a physical attribute, like the relative pose between parts $i$ and $j$, or whether parts $i$ and $j$ are on the same rigid body. On the other hand, $a_{ij}\parens{s_m}$ predicts the value of that same attribute under the hypothesis that the structure is in state $s_m$. $g(\cdot, \cdot)$ is a comparison function that gives higher values to better matches, such as the inner product or the inverse Euclidean distance.


\subsubsection{Scoring assembly sequences}
The total score of a sequence is made up of the scores of each segment, along with
sequence features $f_{seq}$ which model the relationship between assembly actions and kinematic structure:
\begin{align} \label{eq:score}
\begin{split}
    score \parens{y, s} = \sum_{m=1}^{M}
        &score \parens{y_{b_m:e_m}, s_m, s_{m+1}} \\
        + &W_{seq}^T f_{seq} \parens{s_m, s_{m + 1}}.
\end{split}
\end{align}
In Eq. \ref{eq:score}, $b_m$ is the beginning index of segment $m$, $e_m$ is the ending index, and $y_{b_m:e_m} = y_{b_m}, \ldots, y_{e_m}$ is the part of the observed sequence corresponding to that segment.
For our sequence features, we use empirical log transition probabilities estimated from a set of training data.

%
%


\subsection{Segmenting and classifying assembly sequences}
To recognize assembly actions we need to find $s^*(y)$, the best labeling of assembly structures for a particular observation sequence:
\begin{align} \label{eq:decode}
\begin{split}
    s^*(y) = \argmax_{s \in \mathcal{S}^{\otimes T}} \sum_{m=1}^{M}
         &\sum_{t=b_m}^{e_m} W_{obs}^T f_{obs} \parens{y_t, s_m, s_{m + 1}} \\
            + &W_{seq}^T f_{seq} \parens{s_m, s_{m + 1}}.
\end{split}
\end{align}

In some cases segment boundaries for these labels are already known---for example, they may be provided by an upstream system.
In this case, the Viterbi algorithm solves Eq. \ref{eq:decode} in $O(|\mathcal{S}|^2 T)$ time, where $\mathcal{S}$ represents the vocabulary of spatial assemblies.
If boundaries are not known \emph{a priori}, the sequence can be jointly segmented and classified using a segmental generalization of the Viterbi algorithm with worst-case runtime complexity $O(|\mathcal{S}|^2 T^2)$.
The quadratic dependence in the sequence length can be reduced to a constant factor $O(k |\mathcal{S}| T)$ if there is an upper bound on the duration of a segment \cite{sarawagi-cohen-2004} or the number of segments in a sequence \cite{lea_SegCNN}.

\section{Experiments}
\label{sec:experiments}
In this section we present two experiments evaluating the performance of our method
on datasets that were collected during studies targeted at human-robot interaction or spatial cognition. As yet, no public benchmarks exist for testing assembly action recognition methods in the general case. However, the datasets we consider illustrate two realistic application scenarios.

\subsection{IKEA furniture}
\label{sec:ikea}
We first apply our model to the furniture-assembly demonstration dataset of Wang, Ajaykumar, and Huang \cite{robots-wang-hri-2020}. In this dataset, 12 participants demonstrated the assembly of an IKEA chair in two different conditions, resulting in a total of 24 videos. The authors made 18 of these videos available to us (corresponding to both conditions for 9 of the 12 participants). In the first condition participants were asked to assemble the structure as efficiently as possible, and in the second they were asked to demonstrate its assembly ``like a YouTuber".
The chair consists of six different parts: the left and right sides, two support beams, a seat, and a backrest. The left and right sides have three contact points (screw holes), while the rest of the parts have two. In Figure \ref{fig:blocks-frames} we show the component parts laid out before assembly, and the finished chair after assembly.
In these trials first-person video was collected from two Logitech C930 cameras which were stacked vertically and mounted on the participant's head. To facilitate easier tracking, every furniture part was also marked with several AR tags.

\begin{figure}[t]
    \centering
    \includegraphics[scale=0.19]{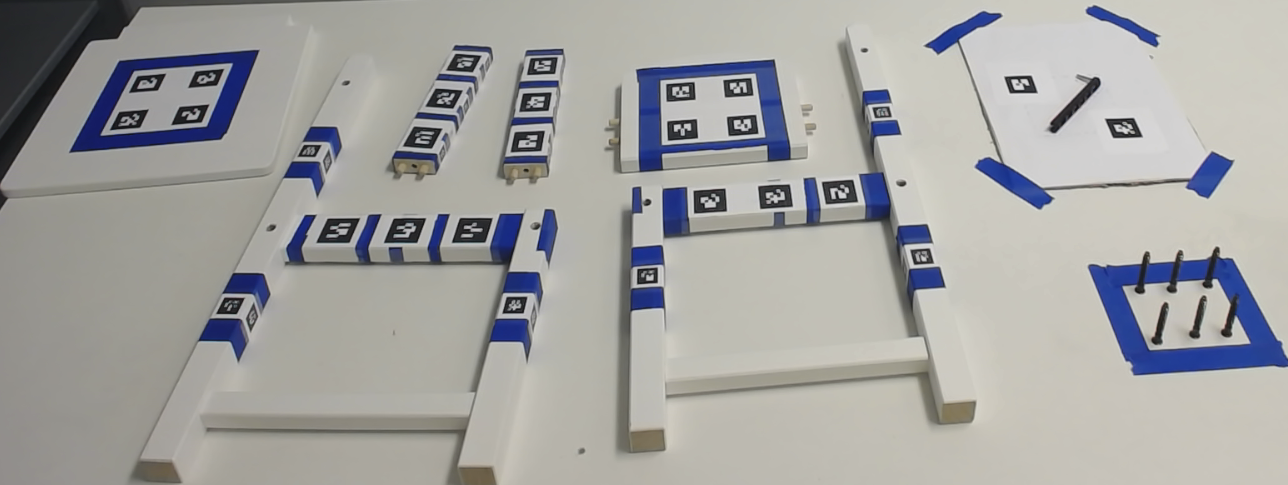}
    
    \vspace{0.25cm}
    
    \includegraphics[scale=0.188]{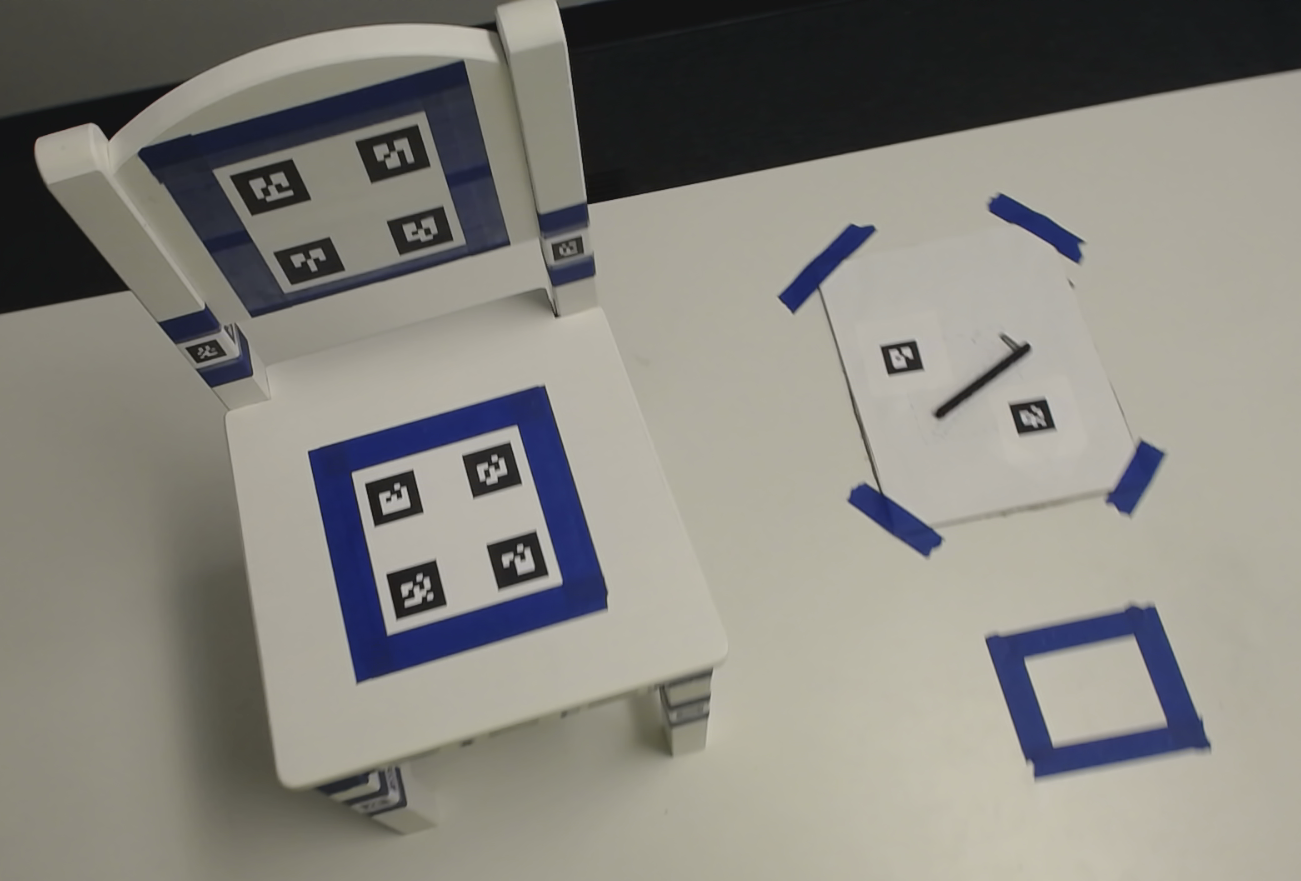}
    \caption{
        TOP: IKEA parts before assembly.
        BOTTOM: Assembled IKEA chair.
    }
    \label{fig:ikea-chair}
\end{figure}

We labeled each video with the assembly actions that happen in it.
For ease of implementation, the partial kinematic graph $\delta s$ that defines an action is annotated one edge at a time.
All labels have the following format:
\begin{equation*}
    \mathtt{action(object_1, object_2, contact ~ points)}. 
\end{equation*}
The \texttt{action} can be one of \texttt{connect} or \texttt{disconnect}.
Each of the \texttt{object} arguments identifies one of the six available parts in the setup.
The contact points identify which screw holes in each part are connected to each other.

We parse these action sequences to produce a corresponding sequence of assembly configurations. Because the beams of the chair are symmetric and identical, some assemblies are indistinguishable from each other. When parsing, we consider assemblies and actions equivalent up to a swap in the placements of the two beams or a 180-degree rotation of a beam in the X-Y plane. In total, there are 10 unique assembly structures in the dataset.

\subsubsection{Observation features}
We obtain an estimated part pose from each bundle of AR markers and each camera using the ALVAR augmented-reality software library.
We then average these pose estimates to obtain a single pose for each part: we compute the Euclidean mean for the position, and the L2 chordal mean for the orientation \cite{Hartley2013}.

Using the estimated part poses, we compute the translation between each pair:
\begin{align}
    \hat{a}_{ij}(I) = \hat{x}_i(I) - \hat{x}_j(I) = \delta \hat x_{ij}(I)
\end{align}
Similarly, for each hypothesis assembly structure $s$, we predict the translation between each pair of parts:
\begin{align}
    a_{ij}(s) = x_i(s) - x_j(s) = \delta x_{ij}(s)
\end{align}

We compare estimated part transforms with predicted ones using the Euclidean distance (scaled by a constant factor $\lambda$). If a pair of parts is not connected in the hypothesis assembly structure $s$, we impose a constant penalty $\alpha$ (this prevents the model from preferring assembly structures with more degrees of freedom):
\begin{align} \label{eq:ikea-obs-feats}
\begin{split}
    f_{obs} \parens{I_t, s_m} = - \sum_{i=1}^{|\mathcal{V}|} \sum_{j=1}^{i-1}
        &\lambda \norm{ \delta \hat{x}_{ij}\parens{I_t} - \delta x_{ij} \parens{s_m}} E_{ij}(s) \\
        + &\alpha (1 - E_{ij}(s)).
\end{split}
\end{align}
In Eq. \ref{eq:ikea-obs-feats}, $E_{ij}(s_m)$ represents element $i, j$ of $s_m$'s adjacency matrix.

\subsubsection{Experimental Setup}
\label{sec:experiments_ikea_setup}
We use a leave-one-video-out cross-validation setup. For each fold, we estimate the log transition probabilities, along with parameters $\alpha$ and $\lambda$, on the 17 training videos, and evaluate performance on the single held-out test video.
We report frame-level accuracy and edit score
averaged across folds, along with their standard deviations.
Frame-level accuracy measures the proportion of samples that were classified correctly, and evaluates the system's overall performance in the joint segmentation and classification task. The edit score is defined in \cite{lea-icra-2016} as $1 - d(s, s') / \max\{|s|, |s'|\}$, where $d(s, s')$ is the Levenshtein edit distance.
This metric evaluates performance at the segment level: it penalizes sequences with mis-classified or mis-ordered segments, but not sequences whose segment boundaries do not overlap. 

\subsubsection{Results}
\label{sec:experiments_ikea_results}
\begin{table}[ht]
	\centering
    \caption{IKEA furniture-assembly results}
    \label{tab:results-ikea}
    \renewcommand{\arraystretch}{1.5}


\begin{filecontents*}{tables/ikea-results.csv}
name,Accuracy,Edit Score,Overlap Score
assembly,69.0+/-13.3,94.0+/-9.0,51.8+/-14.1
action,70.8+/-11.2,89.9+/-15.2,52.7+/-13.2
\end{filecontents*}
\pgfplotstabletypeset[
    col sep=comma,
    columns/name/.style={string type, column type=r, column name={}},
    every head row/.style={before row=\toprule,after row=\midrule},
    every last row/.style={after row=\bottomrule},
    every column/.style={string type, string replace*={+/-}{$\pm$}},
    columns={[index]0, [index]1, [index]2}
]{tables/ikea-results.csv}
\end{table}
\begin{figure*}[t]
    \centering
    \includegraphics[scale=0.6]{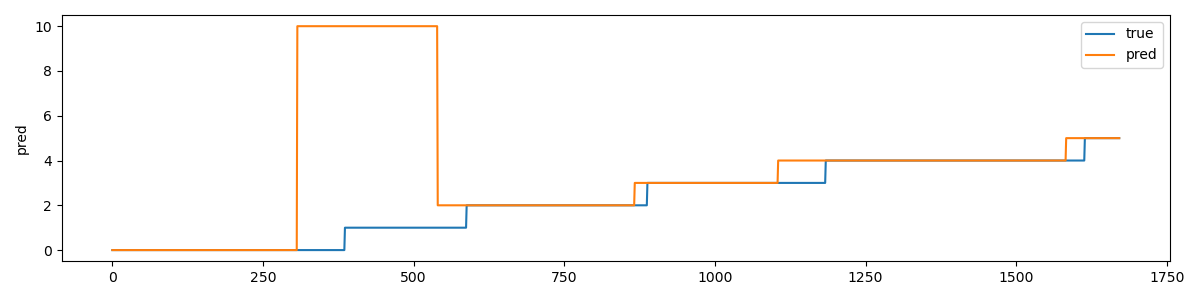}
    \caption{
        In this example, the model confuses the order in which beam parts were connected to the left side of the chair.
        The system's performance on this sequence is close to is average performance: accuracy is  74.8\% and edit score is 83.3\%.
    }
    \label{fig:ikea-eg}
\end{figure*}

In Table \ref{tab:results-ikea} we report metrics evaluated on both assembly sequences and action sequences. We compute action sequences by taking the difference of consecutive assembly segments. The system's average assembly edit score of 94\% shows that it almost always retrieves the correct assembly sequence. This is due in part to the sequence model and the constrained nature of the dataset---errors that might occur if the observation model were operating independently (\eg due to part occlusion or tracking errors) get corrected when when we require that each assembly action must be consistent with the last. The average framewise accuracy of the system is around 70\%, which shows that the system is not operating solely on sequence-model information. When our system makes an error, it usually misclassifies an action near the beginning of a sequence or makes small mistakes in identifying segment boundaries. In Figure \ref{fig:ikea-eg} we present an example sequence exhibiting both of these error types.

\subsection{Toy Blocks}
\label{sec:blocks}
This dataset is based on the construction of six abstract DUPLO assemblies, and exhibits some interesting challenges in computer vision that have not yet been solved:
the block model is frequently occluded during the build process,
the assembly and parts can be moved around arbitrarily in the frame,
extremely high precision ($<$15mm) is necessary
in order to determine the exact nature of each connection
(\ie which studs are joined),
there is much more variability in the attested assemblies (more than 300 unique assemblies result from 145 videos).

\begin{figure*}[t]
    \centering
    \includegraphics[scale=0.2]{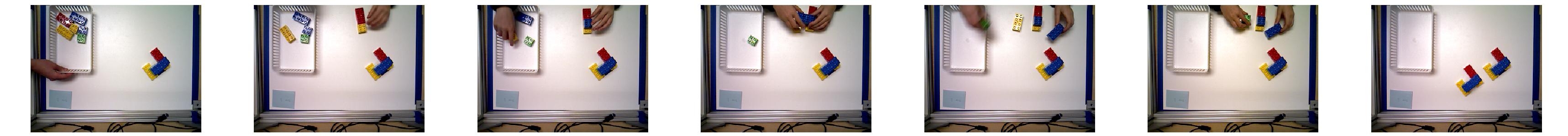}
    \caption{
        Example keyframes from a block assembly video. Each image represents the clearest view of a block assembly in the video.
    }
    \label{fig:blocks-frames}
\end{figure*}

During each of the 145 trials, we recorded RGB and depth video at a resolution of 240x320 pixels and frame rate of 30 Hz from a Primesense Carmine camera mounted in an overhead position.
Each time a reference model was placed in the lower right-hand corner of the frame,
and a basket of blocks is placed on the left-hand side.
In Figure \ref{fig:blocks-frames} we show a sequence of example images summarizing one assembly video.
In addition, we recorded acceleration and angular velocity signals using wireless inertial measurement units (IMUs) embedded in each block. The IMU signals were sampled at 50 Hz, and every (IMU and RGBD) sample was marked with the global time when it was received by the data collection computer.

We labeled these videos using the same format as the IKEA dataset. In total, there are 311 unique assemblies in the dataset.
Because the block model is usually completely occluded during a manipulation action,
it is crucial to provide the parser with high-quality views.
For a subset of 61 videos, we manually label the least-occluded image for each segment.
In this subset, there are 132 unique assemblies.


\subsubsection{Observation features}
Our method builds on \cite{blocks-jones-wacv-2019}, which developed a system for recognizing assemblies from videos alone by rendering template images of spatial assemblies and registering them to observed video frames.
This method was precise, but not accurate: the visual modality captures the complete scene, which contains all the fine-grained detail necessary to disambiguate different connections between the same pairs of parts. However, the scene also contains noise, like distracting and occluding objects, and there are several opportunities for upstream failures in the processing chain.
We augment this video model with features from the inertial data source, in the belief that these modalities should complement each other well.
The inertial modality is accurate, but not precise: our inertial signals are not subject to noise or occlusion, but they can only determine if pairs of blocks are moving on the same rigid body. However, including information from the visual modality should resolve the inertial model's ambiguities.

\textbf{Image features}
Following \cite{blocks-jones-wacv-2019}, we use a pixel-level observation model for our video features.
We render an image of the hypothesis, then register it to the segment using a sum of square errors objective.
%
%
We use the score of the best registration to represent
an image's compatibility with a given assembly hypothesis:
\begin{align}
\label{eq:obs-blocks}
    f_{obs} \parens{I_t, s_m}
        &= \max_{x \in \mathcal{X}(s_m)} - \sum_{(r, c) \notin \mathcal{M}(I_t)}
            \norm{ I_t(r, c) - T(r, c; x) } ^ 2,
\end{align}
In Eq. \ref{eq:obs-blocks}, $\mathcal{X}(s_m)$ is the set of object poses consistent with the constraints imposed by assembly structure $s_m$, $T( \cdot; x)$ is a rendered template image corresponding to object poses $x$, and $\mathcal{M}(I_t)$ is a mask that identifies skin pixels in the image. We ignore skin pixels in the registration score because they can occlude the block models.


\textbf{Inertial features}
We incorporate information from the IMU signals by predicting which blocks are connected (\ie on the same rigid body).
We represent this quantity using an indicator variable $c_{ij}$. 
Its value is 1 if parts $i$  and $j$ are on the same rigid body, and 0 if they are not.
Using a learned classifier, we estimate the value of $c_{ij}$ from observed angular velocity measurements $\omega$:
\begin{align}
    \hat{a}_{ij}(\omega) = \CondProb{c_{ij}}{\omega_i, \omega_j}
\end{align}
We also predict its value for a hypothesized structure $s$:
\begin{align}
    a_{ij}(s) = \CondProb{c_{ij}}{s}
\end{align}
$c_{ij,t}$ can be determined from the assembly structure $s$ by checking whether object $i$ and object $j$ are in the same connected component (\ie whether a path exists connecting vertex $i$ and vertex $j$). Thus, $a_{ij}(s)$ is a one-hot vector.

We rescale both attributes so their entries are in the range (-1, 1) instead of (0, 1),
and we compare estimated part transforms with predicted ones using the inner product.
This way, estimated and predicted attributes contribute a positive score if their predictions agree, and contribute a negative score if their predictions do not:
\begin{align}
\begin{split}
    f_{obs} \parens{\omega_t, s_m} = \sum_{i=1}^{|\mathcal{V}|} \sum_{j=1}^{i-1}
        \angles{2 \hat{a}_{ij}(\omega_t) - 1, 2 a_{ij}(s_m) - 1}
\end{split}
\end{align}

\textbf{Connection classification}
We train a temporal convolutional network (TCN) \cite{colin-tcn} to predict $c_{ij}$.
We use 5-fold cross-validation on the all 145 videos of the blocks dataset, and hold out 25\% of the training data to use as validation in each fold.
We train using cross-entropy loss for 15 epochs, and choose the model with highest F1 score on the heldout set. We use the Adam optimizer with a learning rate of 0.001.
Our network has 6 layers with the following number of channels: [8, 8, 16, 16, 32, 32], and kernel size 25. We use a dropout rate of 0.2.
Overall, the performance of this model is quite good: it retrieves block connections with an accuracy of 92.8 $\pm$ 1.5\%, precision 87.1 $\pm$ 3.1\%, recall 93.6 $\pm$ 1.8\%, and F1 score of 90.2 $\pm$ 1.6\%.

\subsubsection{Experimental Setup}
\label{sec:experiments_blocks_setup}
In this experiment, we evaluate our system on the subset of 61 videos with keyframes using leave-one-out cross-validation.
We use ground-truth segments and keyframes, making this a sequential classification task rather than joint segmentation and classification.
For each fold, we estimate the model's state space and state transition parameters using 61 videos and test on the remaining one.
Since videos are pre-segmented in this experiment, we only report the edit score (and not the accuracy).

\subsubsection{Results}
\label{sec:experiments_blocks_results}
\begin{table}[ht]
	\centering
    \caption{Results: Block-building dataset}
    \label{tab:results-blocks}
    \renewcommand{\arraystretch}{1.5}
\begin{filecontents*}{tables/blocks-results.csv}
name,Edit Score (assembly),Edit Score (action)
VIDEO,61.2+/-36.3,58.6+/-35.9
VIDEO + IMU,77.5+/-28.6,72.8+/-29.8
\end{filecontents*}
\pgfplotstabletypeset[
    col sep=comma,
    columns/name/.style={string type, column type=r, column name={ } },
    every head row/.style={before row=\toprule,after row=\midrule},
    every last row/.style={after row=\bottomrule},
    every column/.style={string type, string replace*={+/-}{$\pm$}}
]{tables/blocks-results.csv}
\end{table}
%
%
\begin{table}[ht]
	\centering
    \caption{Results: Ablation study}
    \label{tab:ablation-blocks}
    \renewcommand{\arraystretch}{1.5}
\begin{filecontents*}{tables/blocks-ablation.csv}
name,Edit Score, Rel. Improvement
IMU,40.3+/-18.1,
VIDEO,43.5+/-23.0,5.66
VIDEO + IMU,54.8+/-21.5,25.0
VIDEO + IMU + SEQ,77.5+/-28.6,101
\end{filecontents*}
\pgfplotstabletypeset[
    col sep=comma,
    columns/name/.style={string type, column type=r, column name={ } },
    every head row/.style={before row=\toprule,after row=\midrule},
    every last row/.style={after row=\bottomrule},
    every column/.style={string type, string replace*={+/-}{$\pm$}}
]{tables/blocks-ablation.csv}
\end{table}
\begin{figure*}[t]
    \centering
    \includegraphics[scale=0.62]{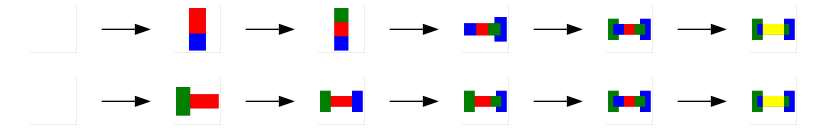}
    \caption{
        Example of an error produced by our system (model orientation is arbitrary in these images). TOP: Predicted sequence. BOTTOM: Ground-truth sequence. The edit distance of 50\%, but the model's incorrect predictions are similar to the ground truth.
    }
    \label{fig:blocks-eg}
\end{figure*}

In Table \ref{tab:results-blocks}, we compare the performance of the baseline video-only model with the IMU-video fusion system. Including our IMU features improves the average assembly edit score by 16\% absolute, corresponding to a 42\% relative reduction in error rate. 
This performance is especially impressive in light of three observations: first, the system is selecting from more than 130 assemblies at every sample. Second, more than 20\% of the assemblies in this dataset only occur once. Because we estimate the model's vocabulary and transition probabilities from the assemblies attested in the training set, this imposes an upper bound of around 80\% on our system performance.
Finally, system errors tend to be similar to the true assembly in appearance and shape (in Figure \ref{fig:blocks-eg} we visualize an example).

In Table \ref{tab:ablation-blocks} we show the results of an ablation experiment examining the contribution of each feature in our model. The first column is the assembly edit score, and the second column is the relative improvement in error rate (\ie one minus the edit score, which is the normalized edit distance).
The inertial and visual modalities operate more-or-less equally alone at roughly 40\% accuracy. Fusing modalities improves the average error rate by 25\% relative to the visual modality alone. Finally, decoding these fused scores with our sequence model gives an additional 100\% improvement. Taking these results together with those of Table \ref{tab:results-blocks}, we can conclude that all three sources of information contribute to the overall performance of our system.

\section{Conclusion}
\label{sec:conclusion}
In this paper we extended the fine-grained activity recognition setting to address the task of assembly action recognition in its full generality by relating assembly actions and kinematic structures.
Then, we outlined a general method for recognizing assembly actions from observation sequences, along with features that take advantage of a spatial assembly's special structure.
Finally, we evaluated our method empirically on two application-driven data sources:
Applied to an existing IKEA furniture dataset, our system recognizes assembly actions with an average framewise accuracy of 70\%, and an average edit score of 90\%.
Evaluated on a block-building dataset that requires fine-grained geometric reasoning
to distinguish between assemblies, our system attains an edit score of 77\%.

This work opens up interesting directions for future research.
Representing all possible assembly structures can be both a blessing and a curse: while it allows us to parse assembly actions at the level of detail required for applications, it also causes the vocabulary that we must search through to explode in size. In this work we have sidestepped the issue by limiting our vocabulary to the assemblies that we encounter at training time, but our blocks experiments show that further performance gains will require a way to recognize previously-unseen assemblies without searching exhaustively through all possible assemblies.
Second, developing a robust and application-general observation model for video data could improve the performance and usability of assembly action recognition systems.
This would require learned pose estimation models that require relatively few training sequences and which are also robust to partial occlusions---since 3D models of each part are usually available in assembly scenarios, we believe that simulation-to-real transfer is an intriguing possibility for this line of work.

\section{Acknowledgements}
We are grateful to Chien-Ming Huang and Yeping Wang for providing us with the IKEA furniture dataset, and advising on data processing.

\addtolength{\textheight}{-12cm}



{\small
\bibliographystyle{IEEEtran}
\bibliography{icra2020-jdjones}

\begin{thebibliography}{10}
\providecommand{\url}[1]{#1}
\csname url@rmstyle\endcsname
\providecommand{\newblock}{\relax}
\providecommand{\bibinfo}[2]{#2}
\providecommand\BIBentrySTDinterwordspacing{\spaceskip=0pt\relax}
\providecommand\BIBentryALTinterwordstretchfactor{4}
\providecommand\BIBentryALTinterwordspacing{\spaceskip=\fontdimen2\font plus
\BIBentryALTinterwordstretchfactor\fontdimen3\font minus
  \fontdimen4\font\relax}
\providecommand\BIBforeignlanguage[2]{{%
\expandafter\ifx\csname l@#1\endcsname\relax
\typeout{** WARNING: IEEEtran.bst: No hyphenation pattern has been}%
\typeout{** loaded for the language `#1'. Using the pattern for}%
\typeout{** the default language instead.}%
\else
\language=\csname l@#1\endcsname
\fi
#2}}

\bibitem{action-gupta-uist-2012}
\BIBentryALTinterwordspacing
A.~Gupta, D.~Fox, B.~Curless, and M.~Cohen, ``Duplotrack: A real-time system
  for authoring and guiding duplo block assembly,'' in \emph{Proceedings of the
  25th Annual ACM Symposium on User Interface Software and Technology}, ser.
  UIST '12.\hskip 1em plus 0.5em minus 0.4em\relax New York, NY, USA: ACM,
  2012, pp. 389--402. [Online]. Available:
  \url{http://doi.acm.org/10.1145/2380116.2380167}
\BIBentrySTDinterwordspacing

\bibitem{hadfield-blocks-iros-2018}
J.~{Hadfield}, P.~{Koutras}, N.~{Efthymiou}, G.~{Potamianos}, C.~S.
  {Tzafestas}, and P.~{Maragos}, ``Object assembly guidance in child-robot
  interaction using rgb-d based 3d tracking,'' in \emph{2018 IEEE/RSJ
  International Conference on Intelligent Robots and Systems (IROS)}, 2018, pp.
  347--354.

\bibitem{robots-wang-hri-2020}
\BIBentryALTinterwordspacing
Y.~Wang, G.~Ajaykumar, and C.-M. Huang, ``See what i see: Enabling user-centric
  robotic assistance using first-person demonstrations,'' in \emph{Proceedings
  of the 2020 ACM/IEEE International Conference on Human-Robot Interaction},
  ser. HRI ’20.\hskip 1em plus 0.5em minus 0.4em\relax New York, NY, USA:
  Association for Computing Machinery, 2020, p. 639–648. [Online]. Available:
  \url{https://doi.org/10.1145/3319502.3374820}
\BIBentrySTDinterwordspacing

\bibitem{blocks-jones-wacv-2019}
J.~{Jones}, G.~D. {Hager}, and S.~{Khudanpur}, ``Toward computer vision systems
  that understand real-world assembly processes,'' in \emph{2019 IEEE Winter
  Conference on Applications of Computer Vision (WACV)}, Jan 2019, pp.
  426--434.

\bibitem{costar-blocks}
A.~{Hundt}, V.~{Jain}, C.~{Lin}, C.~{Paxton}, and G.~D. {Hager}, ``The costar
  block stacking dataset: Learning with workspace constraints,'' in \emph{2019
  IEEE/RSJ International Conference on Intelligent Robots and Systems (IROS)},
  2019, pp. 1797--1804.

\bibitem{action-summers-iros-2012}
D.~{Summers-Stay}, C.~L. {Teo}, Y.~{Yang}, C.~{Fermüller}, and Y.~{Aloimonos},
  ``Using a minimal action grammar for activity understanding in the real
  world,'' in \emph{2012 IEEE/RSJ International Conference on Intelligent
  Robots and Systems}, Oct 2012, pp. 4104--4111.

\bibitem{action-yang-acl-2015}
\BIBentryALTinterwordspacing
Y.~Yang, Y.~Aloimonos, C.~Ferm{\"u}ller, and E.~E. Aksoy, ``Learning the
  semantics of manipulation action,'' in \emph{Proceedings of the 53rd Annual
  Meeting of the Association for Computational Linguistics and the 7th
  International Joint Conference on Natural Language Processing (Volume 1: Long
  Papers)}.\hskip 1em plus 0.5em minus 0.4em\relax Beijing, China: Association
  for Computational Linguistics, July 2015, pp. 676--686. [Online]. Available:
  \url{https://www.aclweb.org/anthology/P15-1066}
\BIBentrySTDinterwordspacing

\bibitem{action-vo-cviu-2016}
N.~N. Vo and A.~F. Bobick, ``Sequential interval network for parsing complex
  structured activity,'' \emph{Computer Vision and Image Understanding}, vol.
  143, pp. 147 -- 158, 2016, inference and Learning of Graphical Models\:
  Theory and Applications in Computer Vision and Image Analysis.

\bibitem{kinem-models-costeira-ijcv-1998}
\BIBentryALTinterwordspacing
J.~P. Costeira and T.~Kanade, ``A multibody factorization method for
  independently moving objects,'' \emph{International Journal of Computer
  Vision}, vol.~29, no.~3, pp. 159--179, Sep 1998. [Online]. Available:
  \url{https://doi.org/10.1023/A:1008000628999}
\BIBentrySTDinterwordspacing

\bibitem{kinem-models-sturm-jair-2011}
J.~Sturm, C.~Stachniss, and W.~Burgard, ``A probabilistic framework for
  learning kinematic models of articulated objects,'' \emph{J. Artif. Intell.
  Res. (JAIR)}, vol.~41, pp. 477--526, 05 2011.

\bibitem{kinem-models-niekum-icra-2015}
S.~{Niekum}, S.~{Osentoski}, C.~G. {Atkeson}, and A.~G. {Barto}, ``Online
  bayesian changepoint detection for articulated motion models,'' in \emph{2015
  IEEE International Conference on Robotics and Automation (ICRA)}, May 2015,
  pp. 1468--1475.

\bibitem{kinem-models-martin-ijrr-2019}
\BIBentryALTinterwordspacing
R.~Mart{\'i}n-Mart{\'i}n and O.~Brock, ``Coupled recursive estimation for
  online interactive perception of articulated objects,'' \emph{The
  International Journal of Robotics Research}, vol.~0, no.~0, p.
  0278364919848850, 0. [Online]. Available:
  \url{https://doi.org/10.1177/0278364919848850}
\BIBentrySTDinterwordspacing

\bibitem{crfs}
\BIBentryALTinterwordspacing
C.~Sutton and A.~McCallum, ``An introduction to conditional random fields,''
  \emph{Foundations and TrendsÂ® in Machine Learning}, vol.~4, no.~4, pp.
  267--373, 2012. [Online]. Available:
  \url{http://dx.doi.org/10.1561/2200000013}
\BIBentrySTDinterwordspacing

\bibitem{sarawagi-cohen-2004}
\BIBentryALTinterwordspacing
S.~Sarawagi and W.~W. Cohen, ``Semi-markov conditional random fields for
  information extraction,'' in \emph{Advances in Neural Information Processing
  Systems 17}, L.~K. Saul, Y.~Weiss, and L.~Bottou, Eds.\hskip 1em plus 0.5em
  minus 0.4em\relax MIT Press, 2005, pp. 1185--1192. [Online]. Available:
  \url{http://papers.nips.cc/paper/2648-semi-markov-conditional-random-fields-for-information-extraction.pdf}
\BIBentrySTDinterwordspacing

\bibitem{lea_SegCNN}
C.~Lea, A.~Reiter, R.~Vidal, and G.~D. Hager, ``Segmental spatiotemporal cnns
  for fine-grained action segmentation,'' in \emph{Computer Vision -- ECCV
  2016}, B.~Leibe, J.~Matas, N.~Sebe, and M.~Welling, Eds.\hskip 1em plus 0.5em
  minus 0.4em\relax Cham: Springer International Publishing, 2016, pp. 36--52.

\bibitem{Hartley2013}
\BIBentryALTinterwordspacing
R.~Hartley, J.~Trumpf, Y.~Dai, and H.~Li, ``Rotation averaging,''
  \emph{International Journal of Computer Vision}, vol. 103, no.~3, pp.
  267--305, Jul 2013. [Online]. Available:
  \url{https://doi.org/10.1007/s11263-012-0601-0}
\BIBentrySTDinterwordspacing

\bibitem{lea-icra-2016}
C.~{Lea}, R.~{Vidal}, and G.~D. {Hager}, ``Learning convolutional action
  primitives for fine-grained action recognition,'' in \emph{2016 IEEE
  International Conference on Robotics and Automation (ICRA)}, 2016, pp.
  1642--1649.

\bibitem{colin-tcn}
C.~Lea, R.~Vidal, A.~Reiter, and G.~D. Hager, ``Temporal convolutional
  networks: A unified approach to action segmentation,'' in \emph{Computer
  Vision -- ECCV 2016 Workshops}, G.~Hua and H.~J{\'e}gou, Eds.\hskip 1em plus
  0.5em minus 0.4em\relax Cham: Springer International Publishing, 2016, pp.
  47--54.

\end{thebibliography}
}

\end{document}